\documentclass[11pt,letterpaper]{article}
\usepackage{ijcnlp2017}
\usepackage{times}
\usepackage{latexsym}

\ijcnlpfinalcopy

\def\ijcnlppaperid{***}

\usepackage{graphicx}
\usepackage{times}
\usepackage{latexsym}
\usepackage[inline]{enumitem}

\usepackage{url}

\title{Learning Phrase Embeddings from Paraphrases with GRUs}

\author{Zhihao Zhou \and Lifu Huang \and Heng Ji \\
	\\ Rensselaer Polytechnic Institute \\
  {\tt \{zhouz5, huangl7, jih\}@rpi.edu}}

\date{}

\begin{document}

\maketitle
\begin{abstract}

Learning phrase representations has been widely explored in many Natural Language Processing (NLP) tasks (e.g., Sentiment Analysis, Machine Translation) and has shown promising improvements. Previous studies either learn non-compositional phrase representations with general word embedding learning techniques or learn compositional phrase representations based on syntactic structures, which either require huge amounts of human annotations or cannot be easily generalized to all phrases. In this work, we propose to take advantage of large-scaled paraphrase database and present a pair-wise gated recurrent units (pairwise-GRU) framework to generate compositional phrase representations. Our framework can be re-used to generate representations for any phrases. Experimental results show that our framework achieves state-of-the-art results on several phrase similarity tasks.



\end{abstract}
\section{Introduction}
Continuous vector representations of words, also known as word embeddings, have been used as features for all kinds of NLP tasks such as Information Extraction~\cite{lample2016neural,zeng2014relation,feng2016language,huang2016liberal}, Semantic Parsing~\cite{chen2014fast,zhou2015end,konstas2017neural}, Sentiment Analysis~\cite{socher2013recursive,kalchbrenner2014convolutional,kim2014convolutional,tai2015improved}, Question Answering~\cite{tellex2003quantitative,kumar2015ask} and machine translation~\cite{cho2014learning,zhang2014bilingually} and have yielded state-of-the-art results. However, single word embeddings are not enough to express natural languages. In many applications, we need embeddings for phrases. For example, in Information Extraction, we need representations for multi-word entity mentions, and in Question Answering, we may need representations for even longer question and answer phrases. 

Generally, there are two types of models to learn phrase emmbeddings: noncompositional models and compositional models. Noncompositional models treat phrases as single information units while ignoring their components and structures. 
Embeddings of phrases can thus be learned with general word embedding learning techniques~\cite{mikolov2013distributed,yin2014exploration,yazdani2015learning}, however, such methods are not scalable to all English phrases and suffer from data sparsity.

On the other hand, compositional models derives a phrase's embedding from the embeddings of its component words~\cite{socher2012semantic, mikolov2013distributed,yu2015learning,poliak2017efficient}. Previous work have shown good results from compositional models which simply used predefined functions such as element-wise addition~\cite{mikolov2013distributed}. However, such methods ignore word orders and cannot capture complex linguistic phenomena. Other studies on compositional models learn complex composition functions from data. For instance, the Recursive Neural Network~\cite{socher2012semantic} finds all linguistically plausible phrases in a sentence and recursively compose phrase embedding from subphrase embeddings with learned matrix/tensor transformations. 

Since compositional models can derive embeddings for unseen phrases from word embeddings, they suffer less from data sparsity. However, the difficulty of training such complex compositional models lies in the choice of training data. Although compositional models can be trained unsupervisedly with auto encoders such as the Recursive Auto Encoder~\cite{socher2011semi}, such models ignore contexts and actual usages of phrases and thus cannot fully capture the semantics of phrases. Some previous work train compositional models for a specific task, such as Sentiment Analysis~\cite{socher2013recursive,kalchbrenner2014convolutional,kim2014convolutional} or syntactic parsing~\cite{socher2010learning}. But these methods require large amounts of human annotated data. Moreover, the embeddings obtained will be biased to a specific task and thus will not be applicable for other tasks. A more general source of training data which does not require human annotation is plain text through language modeling. For example, ~\newcite{yu2015learning} trained compositional models on bigram noun phrases with the language modeling objective. However, using the language modeling objective to train compositional models to compose every phrase in plain text would be impractical for large corpus. 


In this work, we are aiming to tackle these challenges and generate more general and high-quality phrase embeddings. While it's impossible to provide ``gold'' annotation for the semantics of a phrase, we propose to take advantage of the large-scaled paraphrases, since the only criteria of determining two phrases are parallel is that they express the same meaning. This property can be naturally used as a training objective. 

Considering this, we propose a general framework to train phrase embeddings on paraphrases. We designed a pairwise-GRU architecture, which consists of a pair of GRU encoders on two paraphrases. Our framework has much better generalizability. Although in this work, we only trained and tested our framework on short paraphrases, our model can be further applied to any longer phrases. We demonstrate the effectiveness of our framework on various phrase similarity tasks. Results show that our model can achieve state-of-the-art performance on capturing semantics of phrases.

\section{Approach}
\label{2}

In this section, we first introduce a 
large-scaled paraphrase database, the ParaPhrase DataBase (PPDB). Then, we show the basic GRU encoder and our pairwise-GRU based neural architecture. Finally, we provide the training details.

\subsection{Paraphrase Database}

PPDB ~\cite{ganitkevitch2013ppdb} is a database which contains hundreds of millions of English paraphrase pairs extracted from bilingual parallel corpora. It is constructed with the bilingual pivoting method~\cite{bannard2005paraphrasing}. Namely if two English phrases are translated to the same foreign phrase, then the two English phrases are considered to be paraphrases. PPDB comes with 6 pre-packaged sizes: S to XXXL~\footnote{http://paraphrase.org/}. In our work, to ensure efficiency and correctness, we only used the smallest and most accurate S package. To generate training data, we filtered out the paraphrases $(p_1,p_2)$ where

\begin{enumerate}[]\itemsep0pt
\fontsize{10.5pt}{13pt}\selectfont
   \item $p_1$ is identical to $p_2$
    \item $p_1$ or $p_2$ contains any non-letter characters except spaces
    \item $p_1$ or $p_2$ contains words which are not contained in our trained word embeddings
    \item $p_1$ and $p_2$ are both single words
  \end{enumerate}
  
After such a filtering step, we obtained a total number of 406,170 paraphrase pairs.

\subsection{GRU Encoder}
Recurrent neural networks have been proved to be very powerful models to encode natural language sequences. Because of the difficulty to train such networks on long sequences, extensions to the RNN architecture such as the long short-term memory (LSTM) ~\cite{hochreiter1997long} and the gated recurrent units (GRU) ~\cite{cho2014learning} have been subsequently designed, which yielded even stronger performances. Gated structures allow models like the LSTM and the GRU to remember and forget inputs based on the gates' judgment of the inputs' importances, which in turn help the neural networks to maintain a more persistent memory. 

The properties of such gated structures also make these models especially suitable for deriving phrase embeddings: for a compositional model to derive phrase embeddings from word embeddings, it is important that the model recognize words in each phrase which have more impact on the meaning of the phrase. For example, the embedding of ``black cat" should be very close to the embedding of ``cat". Thus the model should partially ignore the word ``black" and let the word ``cat" dominate the final phrase embedding. 

In this work, we chose our compositional model to be the GRU, since it was not only faster to train than the LSTM, but also slightly better-performing on our evaluation tasks. Mathematically, the $j$'th activation of the GRU at time step $t$, $h^j_t$, is given by:
$$h^j_t = (1-z^j_t)h^j_{t-1}+z^j_t\widetilde{h}^j_t$$
where $\widetilde{h}^j_t$ is the current candidate activation and $z^j_t$ is an update gate which dictates the extend to which the current activation is influenced by the current candidate activation and to which it maintains previous activation.

The candidate activation is given by:
$$\widetilde{h}^j_t=tanh(Wx_t+U(r_t\odot h_{t-1}))^j$$
where $U$ and $W$ are transformation matrices, $x_t$ is the current input vector, and $r_t$ is a vector of reset gates which controls how much the model forgets the previous activations.

The update gates and reset gates are both calculated based on the previous activations and current inputs:
$$z^j_t=\sigma (W_zx_t+U_z h_{t-1})^j$$
$$r^j_t=\sigma (W_rx_t+U_rh_{t-1})^j$$

Concretely, given the phrase ``black cat", when it reads the word ``cat", the GRU can learn to forget the word ``black" by setting the update gates $z_t$ close to $1$ and setting the reset gates $r_t$ close to $0$. In this way the final phrase representation of the phrase will mostly be influenced by the word ``cat".

\subsection{PGRU: Pairwise-GRU}
\begin{figure*}[h]
\centering
\includegraphics[width=0.9\textwidth]{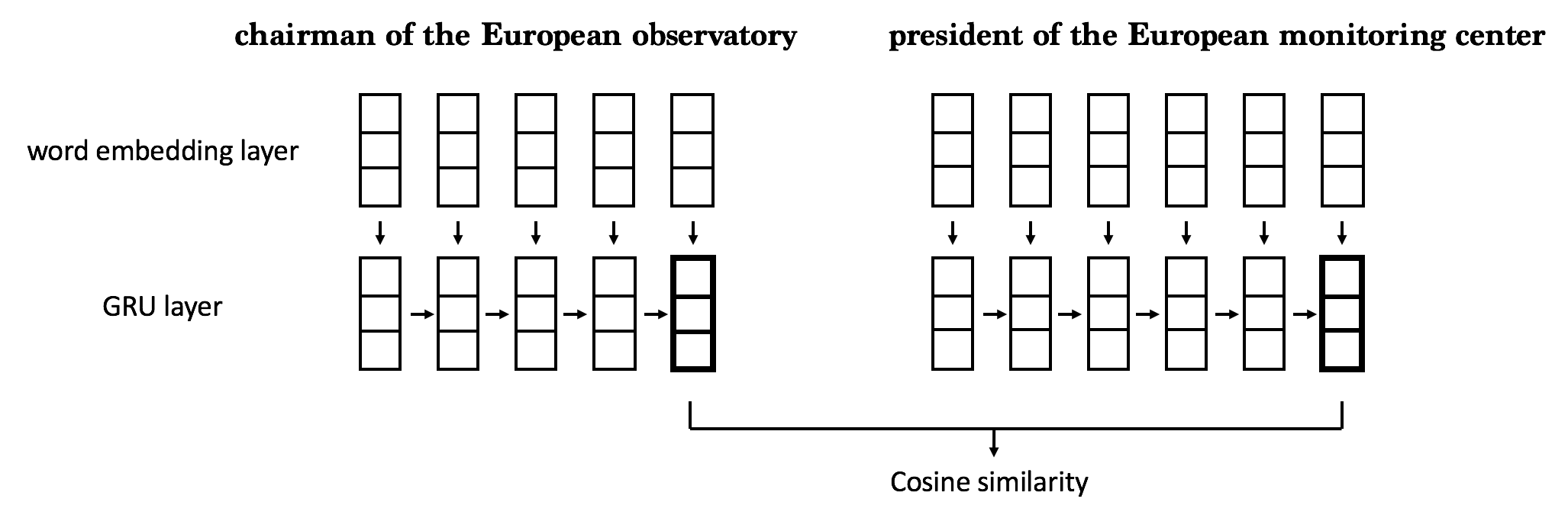}
\caption{PGRU encodes phrase pairs with two GRUs which share the same parameters regardless of phrase lengths. Similarity is calculated by multiplying the two last hidden states with dot product.}
\label{figure1}
\end{figure*}

In order to train GRUs on paraphrases, we propose a Pairwise-GRU (\textit{PGRU}) architecture, which contains two GRUs sharing the same weights, to encode each phrase in the paraphrase pair. Figure~\ref{figure1} shows the overview of our framework. Given a phrase pair $(p_1,p_2)$, e.g., $p_1 = $``\textit{chairman of the European observatory}'' and $p_2 =$``\textit{president of the European monitoring center}'', we first initialize each token in each phrase with a pre-trained word embedding, then the two sequence of word embeddings are taken as input to two GRUs. We take the last hidden layers of the GRUs as the phrase embeddings of $p_1$ and $p_2$, and measure their similarity using cosine similarity with dot product.


Unlike the Recursive Neural Network (\textit{Tree-RNN}) which maps phrases to the word embedding space~\cite{socher2013recursive}, the PGRU maps every phrase, including single words, to a separate phrase embedding space. This characteristic is very important for training the model on paraphrases. For example, given a paraphrase pair ``America" and ``the United States", the \textit{Tree-RNN} only performs matrix/tensor transformations on the embeddings of ``the United States", and generates a new vector representation which would ideally be close to the embedding of ``America". However, since the embedding of ``America" is kept constant,  transformations on ``the United States" has to be very complex.
On the other hand, the PGRU uses GRUs to encode both ``America" and ``the United States" and make their phrase embeddings to be close to each other. Since neither embedding is aimed to be a predefined vector, the transformations can be much simpler and thus much easier to train.


\subsection{Negative Sampling and Training Objectives}
It is not enough for a model to map paraphrases to similar embeddings, since it is also important that it maps semantically different phrases to different embeddings. Thus we need to train the model to distinguish paraphrases from non-paraphrases. Similar to word embedding learning, we use negative sampling~\cite{mikolov2013distributed} to achieve this learning outcome. For each paraphrase pair $(p_1,p_2)$, we select k contrast phrases $c_1,c_2,...,c_{k}$ uniformly at random from the whole paraphrase database regardless of their frequencies of occurrence in the original corpora. Thus the goal of our model is, given the phrase $p_1$, correctly predict that $p_2$ is a paraphrase of $p_1$ and all contrast phrases $c_1,c_2,...,c_{k}$ are not.



We chose our loss function to be the contrastive max-margin loss~\cite{socher2013reasoning}. The main reasoning behind using this training objective is that while we want the cosine similarity of $p_1$ to its paraphrase $p_2$ to be high, it only has to be higher than the similarity of $p_1$ to any contrast phrase $c_i$ by a certain margin so that the model can make correct predictions. Following ~\newcite{socher2013reasoning}, we set the margin to 1.

The contrastive max-margin loss for each training example is defined as:
$$J_t(\theta)=\sum_{i=1}^{k} max(0,1-p_1^Tp_2+p_1^Tc_i)$$
where $p_1$, $p_2$ and $c_i$ are the embeddings of the paraphrases and contrast phrases respectively. And k is the number of contrast phrases.

And the overall loss is calculated by averaging objectives for all training examples:
$$J(\theta)=\frac{1}{T}\sum_{t=1}^{T}J_t(\theta)$$
where T is the number of training examples.

It is worth noting that, although previous embedding training work has predominantly used the negative sampling objective~\cite{mikolov2013distributed}, the contrastive max-margin loss achieved much superior performances in our experiments.

\subsection{Hyperparameters and Training Details}
We used 200-dimensional word embeddings pre-trained with word2vec~\cite{mikolov2013distributed}. We set the number of hidden units of the GRU cell to 200 while using dropout~\cite{hinton2012improving} with a dropout rate of 0.5 on the GRU cells to prevent overfitting. We also used gradient clipping~\cite{pascanu2013difficulty, graves2013generating} with maximum gradient norm set to 5. Training was accomplished with stochastic gradient descent (SGD) with a learning rate of 0.3, a minibatch size of 128 and a total number of epochs of 150.


\section{Experiments}

\subsection{PPDB experiments}

We randomly split the paraphrase pairs chosen from PPDB (as described in Section 2.1) to $80\%$, $10\%$ and $10\%$ as training, development and test sets. To see how the size of training data affects training results, we experimented training with $1\%$, $10\%$ and $100\%$ of our training set. We also experimented setting the number of contrast phrases k to 9, 29 and 99 for each training set size (which correspond to a 10/30/100 choose 1 task for the model). Finally, we evaluated the models trained under each configuration on our test set, where we set k to 99 and computed the accuracy of the model choosing a phrase's paraphrase among contrast phrases. More formally, for a test example $\{p_1,p_2,c_1,c_2,...,c_k\}$, 
the models were given the phrase $p_1$ and asked to choose its paraphrase $p_2$ from the set $\{p_2,c_1,c_2,...,c_k\}$.


To demonstrate the effectiveness of this training procedure, we also included the performance of the commonly used average encoder (\textit{AVG}) on our test set. \textit{AVG} simply takes the element-wise average of a phrase's component word embeddings as the phrase's embedding.

\begin{figure}[h]
\centering
\includegraphics[width=0.48\textwidth]{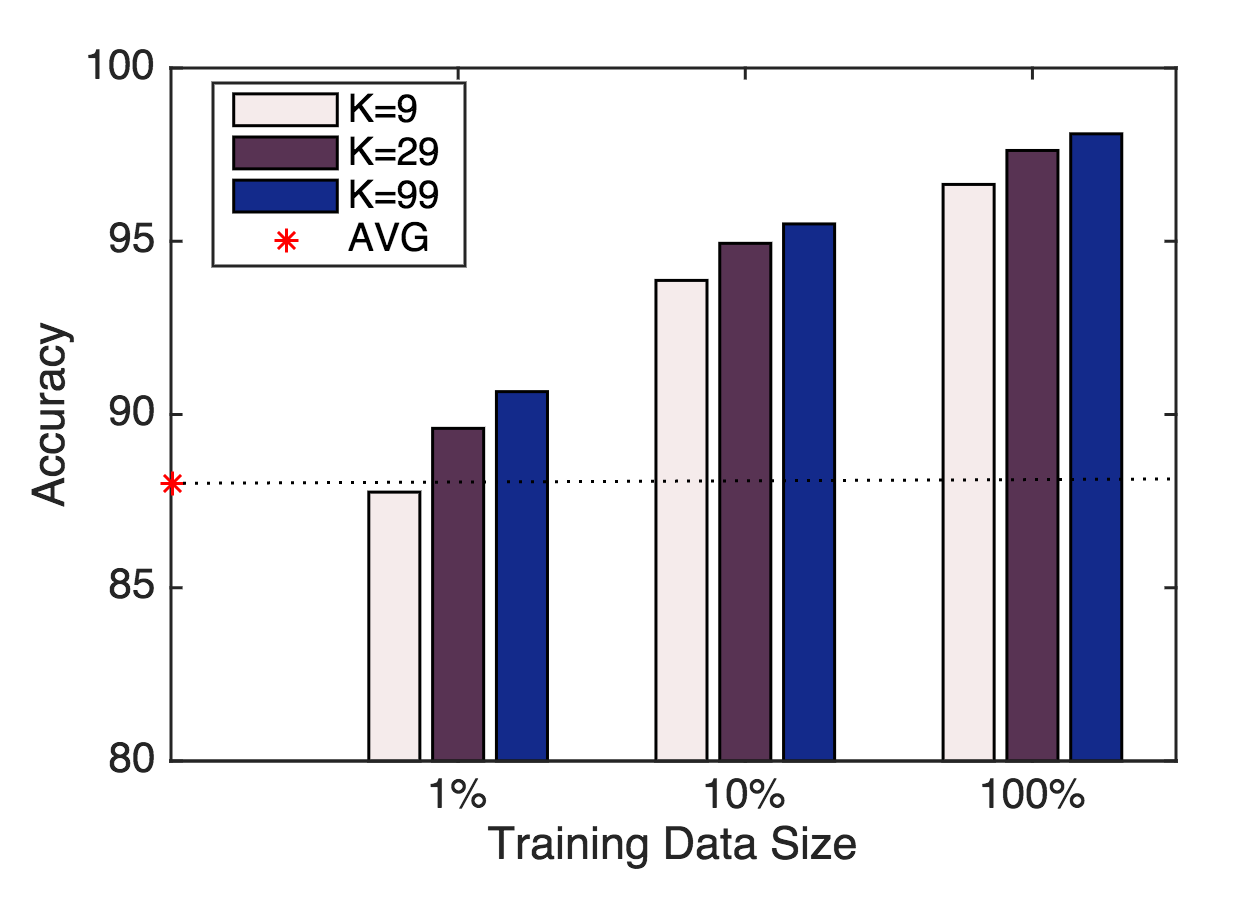}
\caption{Performances of \textit{PGRU} trained under different configurations as well as the performance of \textit{AVG}.}
\label{figure2}
\end{figure}

As shown in Figure~\ref{figure2}, the commonly used \textit{AVG} encoder achieved a score of $88\%$, which suggests that it is indeed a rather effective compositional model. But after adequate training on PPDB, \textit{PGRU} is able to significantly improve upon \textit{AVG}. This shows that \textit{AVG} is not complex enough to fully capture semantics of phrases compared to complex compositional models like the GRU. It also suggests that, during PPDB training, our model can learn useful information about the meaning of phrases which were not learned by word embedding models during word embedding training. From the figure, we can also see consistent performance gain from adding more training data. This again proves that a large paraphrase database is useful for training compositional models. Moreover, for each training set size, while we observe obvious performance gain from increasing k from 9 to 29, the gain from further increasing k to 99 is more moderate. Considering the amount of additional computation required, we conclude that it is not worth the computation efforts to increase k even further.

\subsection{Phrase Similarity Tasks}
\label{3.2}

\begin{table*}
\centering
\begin{tabular}{ |l|c|c|c|c| }
 \hline
 Model & Task Specific & SemEval2013 & Turney2012(5) & Turney2012(10)  \\ \hline
 SUM & False & 65.46 & 39.58 & 19.79  \\
 RAE & False & 51.75 & 22.99 & 14.81  \\
 FCT(LM) & False & 67.22 & \textbf{42.59} & \textbf{27.55}  \\
 GRU(PPDB) & False & \textbf{71.29} & 41.44 & 26.37  \\ \hline
 Tree-RNN  & True & 71.50 & 40.95 & 27.20  \\
 FCT & True & 68.84 & 41.90 & 33.80  \\
 GRU & True & \textbf{73.44} & \textbf{48.88} &\textbf{39.23}  \\ \hline
\end{tabular}
\caption{Performances of our models and baselines on \textbf{SemEval2013}, \textbf{Turney2012(5)} and \textbf{Turney2012(10)}. Models are split into task-specific ones and task-unspecific ones for comparison.}
\label{Results}
\end{table*}

\subsection*{Datasets}
Following~\newcite{yu2015learning}, we 
evaluated our model on human annotated datasets including SemEval2013 Task 5(a) (\textbf{SemEval2013})~\cite{korkontzelos2013semeval} and the noun-modifer problem in Turney2012 (\textbf{Turney2012})~\cite{turney2012domain}. \textbf{SemEval2013} is a task to classify a phrase pair as either semantically similar or dissimilar. \textbf{Turney2012(5)} is a task to select the most semantically similar word to the given bigram phrase among 5 candidate words. In order to test the model's sensitivity to word orders, extended from \textbf{Turney2012(5)}, \textbf{Turney2012(10)} reverse the bigram and add it to the original bigram side. Thus the model needs to choose a bigram from these two bigrams and also choose the most semantically similar word from 5 candidates. Examples for these tasks are shown in Table~\ref{Task-Examples}.


\begin{table*}
\centering
\begin{tabular}{ |c|c|c|c| }
 \hline
 Data Set & Input & Output & train/eval size \\ \hline
 SemEval2013 & (bomb, explosive device) & True & 11722/7814\\  
 Turney2012 & air current , \{wind, gas, sedum, sudorific, bag\} & wind & 680/1500\\ \hline
\end{tabular}
\caption{Examples for \textbf{SemEval2013} and \textbf{Turney2012} as well as the number of training and evaluation examples for each task.}
\label{Task-Examples}
\end{table*}

Both tasks include separate training and evaluation sets. Note that although both tasks only contain unigram and bigram noun phrases, our approach of learning phrase embeddings can be applied to n-grams of any kind. 
We tested the performances of the GRU trained on the provided training set for each task (\textit{GRU}) as well as the GRU trained only on the PPDB data (\textit{GRU(PPDB)}), as described in Section~\ref{2}. For task-unspecific training (\textit{GRU(PPDB)}), we used the training set of each task as development set and applied early stopping.

\subsection*{Baselines}
We compare our results against baseline results reported by~\newcite{yu2015learning}. The baseline method \textit{SUM} is the commonly used element-wise addition method~\cite{mitchell2010composition}. \textit{RAE} is the recursive auto encoder~\cite{socher2011semi} which is unsupervisedly trained to compose phrase embeddings such that the resulting phrase embeddings can be used predict the phrase's composing word embeddings. \textit{FCT}~\cite{yu2015learning} is a compositional model which calculates a phrase's embeddings as a per-dimension weighted average of the component word embeddings while taking into consideration linguistic features such as part of speech tags. \textit{FCT(LM)}~\cite{yu2015learning} is the FCT model trained on news corpus with language modeling objective instead of on the provided training sets for each task. \textit{Tree-RNN} is the recursive neural network~\cite{socher2011semi,socher2013recursive} which builds up phrase embeddings from composing word embeddings with matrix transformations while also taking advantage of POS tags and parse tree structures.

We divide our results to comparisons of task-specific models and comparisons of task-unspecific ones, where for task-specific models, we remove scores from ~\newcite{yu2015learning} which require fine-tuning word embeddings since we are only comparing compositional models. For the sake of comparison, we use the same word embeddings used by ~\newcite{yu2015learning}, although better scores can be achieved by using word embeddings of larger vocabulary size.

\subsection*{Results}
As shown in table~\ref{Results}, \textit{GRU} performs the best among all task-specific models in all three tasks, which proves that GRU is a very powerful compositional model and suggests that it is a suitable model to learn compositional phrase embeddings. \textit{GRU}'s much superior performances on \textbf{Turney2012(5)} and \textbf{Turney2012(10)} can also be attributed to the fact that we used the contrastive max-margin loss (as described in Section 2.4) as training objective, which proved to be more effective in our experiments than the negative sampling objective used by ~\newcite{yu2015learning}.

Among task-unspecific models, \textit{GRU(PPDB)} also achieves strong performances. In all three tasks, \textit{GRU(PPDB)} outperforms \textit{SUM}, suggesting that the compositional model learned from PPDB can indeed be used for other domains and tasks. In particular, on \textbf{Turney2012(10)}, \textit{GRU(PPDB)} improves upon \textit{SUM} by a large margin. This is because unlike \textit{SUM}, GRUs can capture the order of words in natural language. It also suggests that on tasks where word order plays an important role, using GRUs trained on PPDB can be more appropriate than using \textit{SUM}. \textit{GRU(PPDB)} also outperforms \textit{FCT(LM)} on \textbf{SemEval2013} and achieves very close performances to \textit{FCT(LM)} on \textbf{Turney2012(5)} and \textbf{Turney2012(5)} despite the fact that \textit{FCT(LM)} is specifically designed and trained to compose noun phrases, which are the only type of phrases present in these three tasks, whereas our model works for all types of phrases. In addition, unlike the FCT, our method of training GRUs on paraphrases do not need any linguistic features produced by parsers which can be prone to errors.


\section{Related Work}


Phrase embeddings can be learned from either compositional or noncompositional models. Noncompositional models learn phrase embeddings by treating phrases as single units while ignoring their components and structures. But such methods are not scalable to all English phrases and suffer from data sparsity.

Compositional models build phrase embeddings from the embeddings of its component words. Previous work has shown that simple predefined composition functions such as element-wise addition~\cite{mikolov2013distributed}
are relatively effective. However, such methods ignore word orders and are thus inadequate to capture complex linguistic phenomena. 

One way to capture word order and other linguistic phenomena is to learn more complex composition functions from data. For instance, Recursive Neural Networks~\cite{socher2011semi,socher2013recursive} recursively compose embeddings of all linguistically plausible phrases in a parse tree with complex matrix/tensor transformation functions. However, models like this are very hard to train. When there are no human-annotations, we can train each phrase embedding to reconstruct the embeddings of it subphrases in the parse tree~\cite{socher2011semi}, but this objective does not capture the meaning of the phrase. When there are human-annotations, for example, if we have annotated sentiment score for each phrase, we can train the embeddings of phrases to predict their sentiment scores. However, in most cases, we do not have so much human-annotated data. Moreover, since these phrase embeddings are only trained to capture sentiment, they cannot be directly applied to other tasks. Our model also falls under this category, but by training our model on a large paraphrase database, we do not need additional human-annotations and the composition functions learned are not restricted to any specific tasks.

There has also been work on integrating annotation features to improve composition. For example, FCT~\cite{yu2015learning} uses annotation features such as POS tags and head word locations as additional features and compose word vectors with element-wise weighted average. While using such features makes sense linguistically, the assumption that phrase embeddings have to be element-wise weighted average of word embeddings is artificial. Also, the annotation features used by such methods might not be accurate due to parser errors.

Finally, our work also share similarity with neural machine translation. For example~\newcite{cho2014learning} showed phrase embeddings can be learned with the RNN Encoder-Decoder from bilingual phrase pairs. Our model differs from their model in that our model only has the encoder part and it relates two phrases in a phrase pair with cosine similarity instead of conditional probability. We also do not only consider true paraphrase pairs but leverage negative sampling to make the model more robust. In addition, our model is trained on English paraphrases instead of bilingual phrase pairs.
\section{Conclusion}
In this paper, we introduced the idea of training complex compositional models for phrase embeddings on paraphrase databases. 
We designed a pairwise-GRU framework to encode each phrase with a GRU encoder. Compared with previous non-compositional and compositional phrase embedding methods, our framework has much better generalizability and can be re-used for any length of phrases. In addition, the experimental results on various phrase similarity tasks showed that our framework can also better capture phrase semantics and achieve state-of-the-art performances.


\section*{Acknowledgments}
We would thank all the reviewers for the valuable suggestions. This project was supported by the DARPA DEFT and U.S. ARL NS-CTA. The views and conclusions contained in this document are those of the authors and should not be interpreted as representing the official policies, either expressed or implied, of the U.S. Government. The U.S. Government is authorized to reproduce and distribute reprints for Government purposes notwithstanding any copyright notation here on.

\bibliography{acl2017}
\bibliographystyle{ijcnlp2017}

\end{document}